\documentclass[letterpaper, 10 pt, conference]{ieeeconf} 

\usepackage{graphics} 
\usepackage{epsfig} 
\usepackage{mathptmx} 
\usepackage{times} 
\usepackage{amsmath} 
\usepackage{amssymb}  
\usepackage{siunitx} 
\usepackage{gensymb} 
\usepackage[colorlinks=true]{hyperref}
\usepackage{booktabs}
\usepackage{multirow}
\usepackage{xcolor}
\usepackage{booktabs}
\let\labelindent\relax
\usepackage{enumitem}

\newcommand{\mypara}[1]{\par\vspace*{0.8mm}\noindent\textbf{\textit{#1}}}
\newcommand{\etal}{\textit{et al}. }
\newcommand{\ie}{\textit{i}.\textit{e}., }
\newcommand{\eg}{\textit{e}.\textit{g}. }
\DeclareMathOperator*{\argmin}{arg\,min}

\title{\LARGE \bf
Form2Fit: \\Learning Shape Priors for Generalizable Assembly from Disassembly
\vspace{-3mm}
}

\author{
Kevin Zakka$^{1, 2}$,
Andy Zeng$^{2}$,
Johnny Lee$^{2}$,
Shuran Song$^{2,3}$
\vspace{1mm}\\
$^{1}$Stanford University\quad
$^{2}$Google\quad
$^{3}$Columbia University
\vspace{-3mm}
}

\begin{document}

\maketitle
\thispagestyle{empty}
\pagestyle{empty}


\begin{abstract}
Is it possible to learn policies for robotic assembly that can generalize to new objects? We explore this idea in the context of the kit assembly task. Since classic methods rely heavily on object pose estimation, they often struggle to generalize to new objects without 3D CAD models or task-specific training data. In this work, we propose to formulate the kit assembly task as a shape matching problem, where the goal is to learn a shape descriptor that establishes geometric correspondences between object surfaces and their target placement locations from visual input. This formulation enables the model to acquire a broader understanding of how shapes and surfaces fit together for assembly -- allowing it to generalize to new objects and kits. To obtain training data for our model, we present a self-supervised data-collection pipeline that obtains ground truth object-to-placement correspondences by disassembling complete kits. Our resulting real-world system, \textit{Form2Fit}, learns effective pick and place strategies for assembling objects into a variety of kits -- achieving 90\% average success rates under different initial conditions (\eg varying object and kit poses), 94\% success under new configurations of multiple kits, and over 86\% success with completely new objects and kits. Code, videos, and supplemental material are available at \href{https://form2fit.github.io}{https://form2fit.github.io}\footnote{ We thank Nick Hynes, Alex Nichol, and Ivan Krasin for fruitful discussions, Adrian Wong, Brandon Hurd, Julian Salazar, and Sean Snyder for hardware support, and Ryan Hickman for valuable managerial support. We are also grateful for hardware and financial support from Google.
}
\end{abstract}


\section{Introduction}

\begin{figure}[t]
\centering
  \vspace{2mm}
  \includegraphics[width=\linewidth]{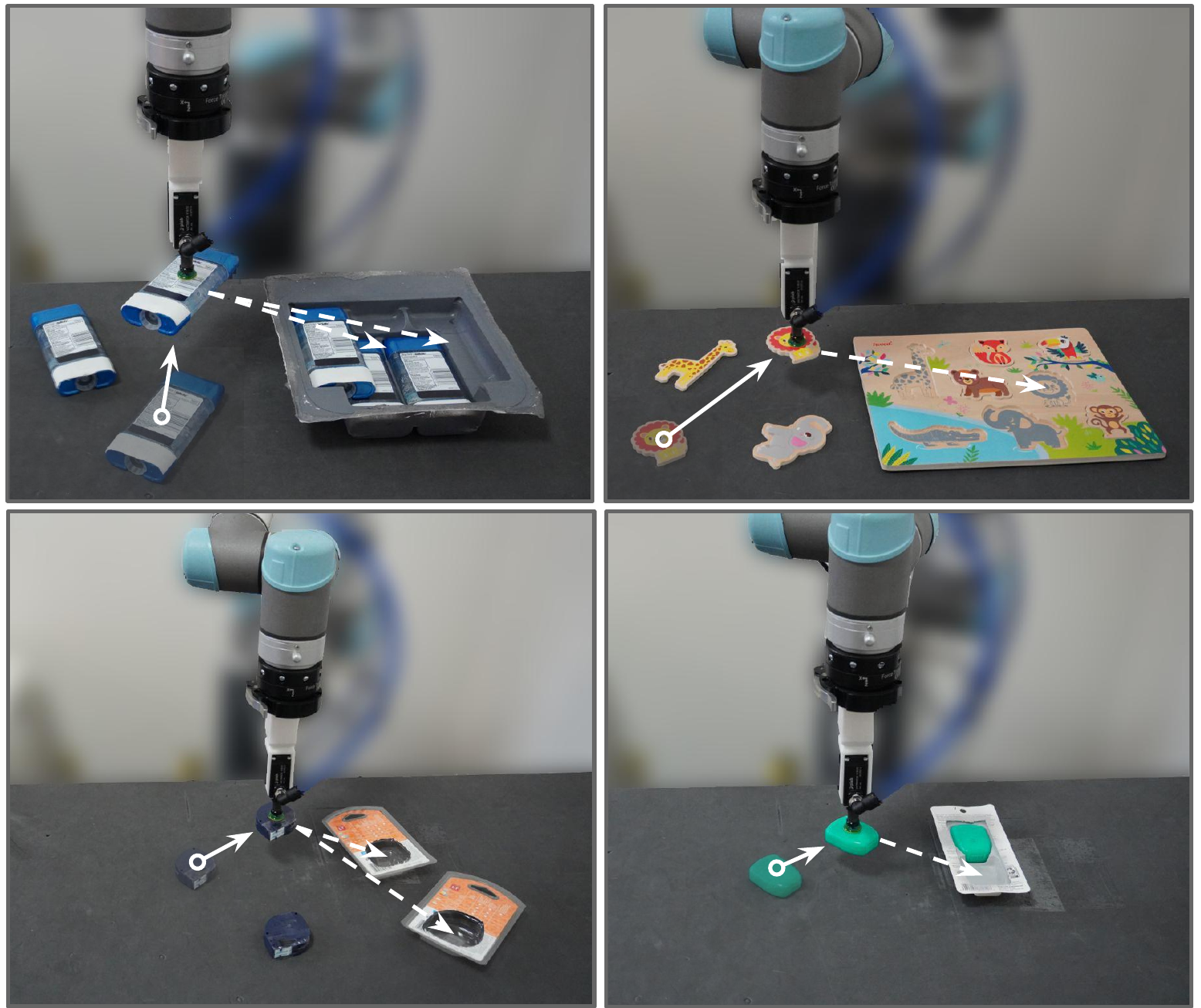}
  \vspace{-5mm}
  \caption{\textbf{Form2Fit} learns to assemble a wide variety of kits by finding geometric correspondences between object surfaces and their target placement locations. By leveraging data-driven shape priors learned from multiple kits during training, the system generalizes to new objects and kits.}
  \label{fig:teaser}
  \vspace{-5mm}
\end{figure}

Across many assembly tasks, the shape of an object can often inform how it should be fitted with other parts. For example, in kit assembly (\ie placing object(s) into a blister pack or corrugated display to form a single unit -- see examples in Fig. \ref{fig:teaser}), the profile of an object likely matches the silhouette of the cavity in the paperboard packaging (\ie kit) that it should be placed into.

While these low-level geometric signals can provide useful cues for both perception and planning, they are often overlooked in many modern assembly methods, which typically abstract visual observations into 6D object poses and then plan on top of the inferred poses. By relying heavily on accurate pose information, these algorithms remain unable to generalize to new objects and kits without task-specific training data and cost functions. As a result, these systems also struggle to quickly scale up to the subclass of real-world assembly lines that may see new kits every two weeks (\eg due to seasonal items and packages).

In this work, we explore the following: if we formulate the kit assembly task as a shape matching problem, is it possible to learn policies that can generalize to new objects and kits? To this end, we propose \textbf{\textit{Form2Fit}}, an end-to-end pick and place formulation for kit assembly that leverages shape priors for generalization. Form2Fit has two key aspects:
\begin{itemize}[leftmargin=*]
    \item \textbf{Shape-driven assembly for generalization}. We establish geometric correspondences between object surfaces and their target placement locations (\eg empty holes/cavities) by learning a fully convolutional network that maps from visual observations of a scene to dense pixel-wise feature descriptors. During training, the descriptors are supervised in a Siamese fashion and regularized so that they are more similar for ground truth object-to-placement correspondences. The key idea is that as the network trains over a variety of objects and target locations across multiple kitting tasks, it acquires a broader understanding of how shapes and surfaces fit together for assembly -- subsequently learning a more generalizable descriptor that is capable of matching new objects and target locations.
    \item \textbf{Learning assembly from disassembly.} We present a self-supervised data-collection pipeline that obtains training data for assembly (\ie ground truth motion trajectories and correspondences between objects and target placements) by disassembling completed kits. Classic methods of obtaining training data for assembly (\eg via human tele-operated demonstrations or scripted policies) are often time-consuming and expensive. However, we show that for kit assembly, it is possible to autonomously acquire large amounts of high-quality assembly data by disassembling kits through trial and error with pick and place, then rewinding the action sequences over time.
\end{itemize}

This enables our system to assemble a wide variety of kits under different initial conditions (\eg different rotations and translations) with accuracies of 90\% (rate at which an object is placed in its target kit with the correct configuration), and generalizes to new settings with mixtures of multiple kits as well as entirely new objects and kits unseen during training.

The primary contribution of this paper is to provide new perspectives on robotic assembly: in particular, we study the extent to which we can achieve generalizable kit assembly by formulating the task as a shape matching problem. We also demonstrate that it is possible to acquire substantial amounts of training data for kit assembly by reversing the action sequences collected from self-supervised disassembly. We provide extensive experiments in real settings to evaluate key components of our system. We also discuss some extensions to our formulation, as well as its practical limitations.


\section{Related Work}
\label{sec:citations}

\mypara{Object pose estimation for assembly.} Classic methods for assembly are often characterized by a perception module that first estimates the 6D object poses of observed parts \cite{braun2016pose,collet2011moped,LINEMOD,wang2019normalized,zeng2016multi}, followed by a planner that then optimizes for picking and placing actions based on the inferred object poses and task-informed goal states. For example, Choi \etal \cite{choi2012voting} introduces a voting-based algorithm to estimate the poses of component parts for assembly from 3D sensor data. Litvak \etal \cite{litvak2018learning} leverages simulated depth images to amass the training data for pose estimation. Jorg \etal \cite{jorg2000flexible} uses a multi-sensory approach with both vision and force-torque sensing to improve the accuracy of engine assembly.

While these methods have seen great success in highly structured environments, they require full knowledge of all object parts (\eg with high-quality 3D object models) and/or substantial task-specific training data and task definitions (which requires manual tuning). This limits their practical applicability to kit assembly lines in retail logistics or manufacturing, which can be exposed to new object and kits as frequently as every two weeks. Handling such high task-level variation requires assembly algorithms that can quickly scale or adapt to new objects, which is the focus of our formulation, Form2Fit.

\mypara{Reinforcement learning for assembly.} Another line of work focuses on learning policies for assembly tasks to replace classic optimization-based or rule-based planners. To simplify the task, these works often assume full state knowledge (\ie object and robot poses). For example Popov \etal \cite{popov2017dataefficient} tackles the task of Lego brick stitching, where the state of the environment including brick positions are provided by the simulation environment. Thomas \etal \cite{thomas2018learning} learns a variety of robotic assembly tasks from CAD models, where the state of each object part is detected using QR codes.

More recent works \cite{levine2016end,nair2018time} eliminate the need for accurate state estimation by learning a policy that directly maps from raw pixel observations to actions through reinforcement learning. However, these end-to-end models require large amounts of training data and remain difficult to generalize to new scenarios (train and test cases are often very similar with the same set of objects). In contrast  our system is able to learn effective assembly policies with a much smaller amount of data (500 disassembly sequences), and generalizes well to different object types and scene configurations (different pose and number of objects) without extensive fine-tuning.

\mypara{Learning shape correspondences.} Learning visual and shape correspondences is a fundamental task in vision and graphics, studied extensively in prior work via descriptors \cite{zeng20173dmatch,florence2018dense,schmidt2016self,johnson1999using,frome2004recognizing,rusu2008aligning}. These descriptors are either designed or trained to match between points of similar geometry across different meshes or from different viewpoints. Hence, rotation invariance is a desired property from these descriptors. In contrast to prior work, our goal is to learn a general matching function between objects and their target placements locations (\ie holes/cavities in the kits) instead of matching between two similar shapes. We also want the descriptor to be ``rotation-sensitive'', so that the shape matching result can inform the actions necessary for successful assembly.

\mypara{Learning from reversing time.} Time-reversal is a classic trick in computer vision for learning information from sequences of visual data (\ie videos). For example, many works \cite{wei2018learning,ramanathan2015learning,fernando2017self,misra2016shuffle} study how predicting the arrow of time or the order of frames in a video can be used to learn useful image representations and feature embeddings. Nair \etal \cite{nair2018time} uses time-reversal as supervision for video prediction to learn effective visual foresight policies.

While many of these methods use time-reversal as a means to extract additional information from the order of frames, we instead use time-reversal as a way to generate correspondence labels from pick and place. We observed that a pick and place sequence for disassembly, when reversed, can serve as a valid sequence for quasi-static assembly. This is not true for all assembly tasks, particularly when more complex dynamics are involved, but this observation holds true for a substantial number of kit assembly tasks. Since it is easier to disassemble than assemble, we leverage time-reversed disassembly sequences (\eg obtained from trial and error) to amass training data for assembly.


\section{Method Overview}

\begin{figure*}[]
\centering
  \includegraphics[width=0.9\textwidth]{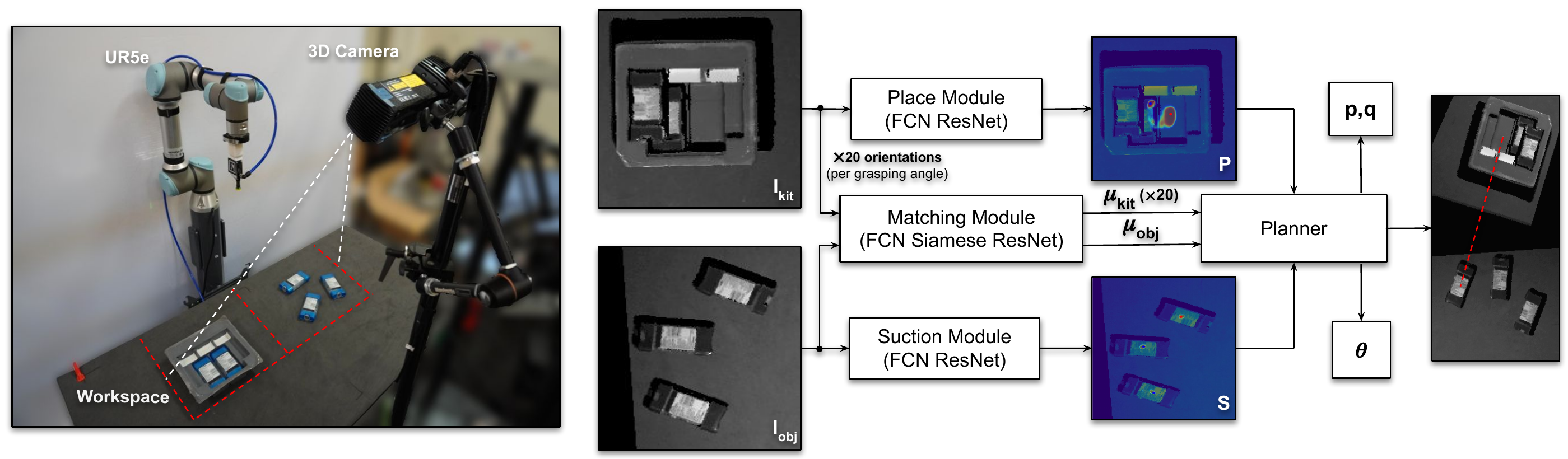}
  \vspace{-1em}
  \caption{\textbf{Overview.} Kit heightmap ($I_\mathrm{kit}$) and object heightmap ($I_\mathrm{obj}$) are generated from a visual observation of the workspace. They are fed to the place and suction modules respectively to produce dense probability maps of place $P$ and suction $S$ success. In parallel, the matching module ingests both kit and object heightmaps to produce kit and object descriptor maps $\mu_\mathrm{kit}$ and $\mu_\mathrm{obj}$ which are fed along with $P$ and $S$ to the planner. The planner integrates this information to produce a picking location $p$, a placing location $q$, and an angle $\theta$ encoding the end-effector rotation about the z-axis.}
  \label{fig:method-overview}
 \vspace{-6mm}
\end{figure*}

Form2Fit takes as input a visual observation $I$ of the workspace (including objects and kits), and outputs a prediction of three parameters: a picking location $p$, a placing location $q$, and an angle $\theta$ that defines a change in orientation between the picking and placing locations. These parameters are used with motion primitives on the robot to execute a respective pick, orient, and place operation. Our learning objective is to optimize our predictions of $p,q,\theta$ such that after each physical execution, an object is thereby correctly placed into its target location and orientation in its corresponding kit. In this work, each prediction of $p,q,\theta$ is \textit{i.i.d.} and conditioned only on the current visual input -- which is sufficient for sequential planning in kit assembly (see Sec. \ref{sec:method-placing} and \ref{sec:method-matching}).

The system consists of three network modules: 1) a suction network that outputs dense pixel-wise predictions of picking success probabilities (\ie affordances) using a suction cup, 2) a placing network that outputs dense pixel-wise predictions of placing success probabilities on the kit, and 3) a matching network that outputs dense, pixel-wise, rotation-sensitive feature descriptors to match between objects and their corresponding placement locations in the kit. We use the placement network directly to infer $q$, then use the suction and matching networks together to infer both $p$ and $\theta$ (see Sec. \ref{sec:method-planning}). Fig. \ref{fig:method-overview} illustrates an overview of our approach.

Our system is trained through self-supervision from time-reversed disassembly: randomly picking with trial and error to disassemble from a fully-assembled kit, then reversing the disassembly sequence to obtain training labels for the suction, placing, and matching networks. In the following subsections, we provide an overview of each module, then describe the details of data collection and training.

\subsection{Visual Representation}
We represent the visual observation of the workspace as a grayscale-depth heightmap image. To compute this heightmap, we capture intensity and depth images from a calibrated and statically-mounted camera, project the data onto a 3D point cloud, and orthographically back-project upwards in the direction of gravity to construct a 2-channel heightmap image representation $I$ with both grayscale (generated from intensity) and height-from-bottom (depth) channels concatenated. Each pixel location thus maps to a 3D position in the robot's workspace. The workspace covers a $1.344\si{\metre}\times0.672\si{\metre}$ tabletop surface. The heightmap has a pixel resolution of $360 \times 464$ with a spatial resolution of $0.002\si {\metre}$ per pixel. We constrain the kit to lie within the left half of the workspace and the objects within the right half. This separation allows us to split the heightmap image into two halves: $I_\mathrm{kit}$ containing the kit, and $I_\mathrm{obj}$ containing the objects, each with a pixel resolution of $360 \times 232$. The supplemental file contains additional details on the representation.

\subsection{Suction Module: Learning Where to Pick}

The suction module uses a deep network that takes as input the heightmap image $I$, and predicts favorable picking locations for a suction primitive on the objects inside the kit (during disassembly) and outside the kit (during assembly). Stable suction points often correspond to flat, nonporous surfaces near an object's center of mass.

\mypara{Suction primitive.} The suction primitive takes as input a 3D location and executes a top-down suction-based grasp centered at that location. This primitive executes in an open loop fashion using stable, collision-free IK solves \cite{diankov_thesis}.

\mypara{Network architecture.} The suction network is a fully-convolutional dilated residual network \cite{he2016deep,long2015fully,yu2017dilated}. It takes as input a grayscale-depth heightmap $I$ and outputs a suction confidence map $S$ with the same size and resolution as that of the input $I$. Each pixel $s_{i}\in{S}$ represents the predicted probability of suction success (\ie suction affordance) when the suction primitive is executed at the 3D surface location (inferred from calibration) of the corresponding pixel $i\in{I}$. The supplemental file contains architectural details.

\vspace{-1mm}
\subsection{Place Module: Learning Ordered Placing\label{sec:method-placing}}

\begin{figure}[t]
\centering
  \includegraphics[width=\linewidth]{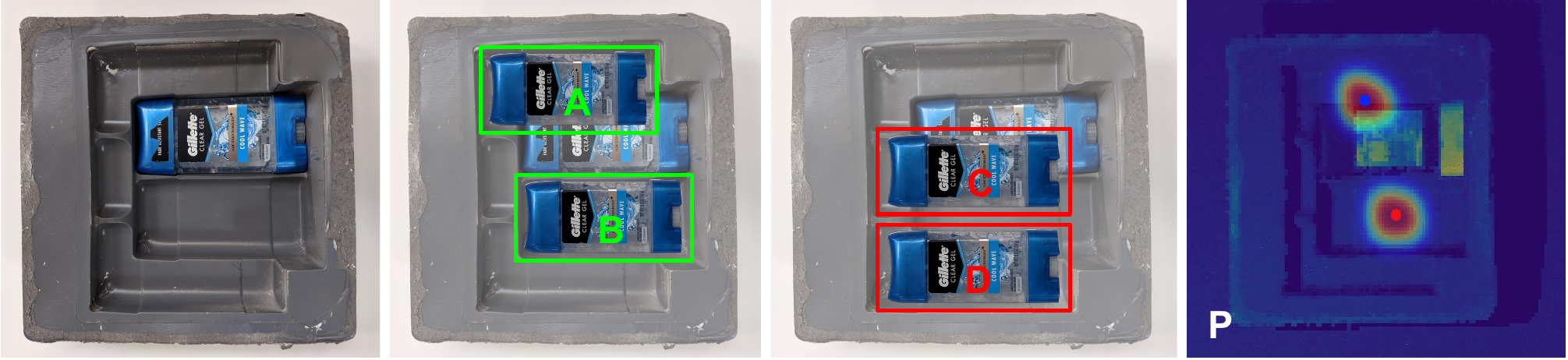}
  \caption{\textbf{An example of ordered assembly} for the deodorant kit. At timestep $t=1$, locations A and B are valid placing positions, while C and D are examples of invalid locations. Our placing module implicitly learns this ordering contraint from its training data (\ie time-reversed disassembly sequences) -- its output prediction P (visualized as a heatmap) shows higher placing confidence values at A and B compared to C and D.}
  \label{fig:place-importance}
  \vspace{-5mm}
\end{figure}

For certain kit assembly tasks, there may be sequence-level constraints that define the order in which objects should be placed into the kit. For example, to successfully assemble a five-pack kit of deodorants (see this test case in Fig. \ref{fig:place-importance}), the bottom layer must be filled with deodorants before the robot can proceed to fill the top layer.

One way our system enables sequential ordering is through a place module that predicts the next best placing location conditioned on the current state (\ie observation) of the environment. The place module consists of a fully-convolutional network (same architecture as suction network) which takes as input the kit heightmap $I_\mathrm{kit}$ and outputs a dense pixel-wise prediction $P$ of placing confidence values over $I$. The 3D locations of the pixels with higher confidence serve as better locations for the suction gripper to approach from a top-down angle while holding an object.

\subsection{Matching Module\label{sec:method-matching}}

While the suction and placing modules provide a list of candidate picking and placing locations, the system requires a third module to 1) \textit{associate} each suction location on the object to a corresponding placing location in the kit and 2) infer the change in object orientation. This matching module serves as the core of our algorithm, which learns dense pixel-wise \textit{orientation-sensitive} correspondences between the objects on the table and their placement locations in the kit.

\mypara{Network architecture.} The matching module consists of a two-stream Siamese network \cite{bromley1994signature}, where each stream is a fully-convolutional residual network with shared weights across streams. Its goal is to learn a function $f$ that maps each pixel of the observed kit and objects in the heightmap $I$ to a $d$-dimensional descriptor space ($d=64$ in our experiments), where closer feature distances indicate better object-to-placement correspondences, \ie $f: I \in \mathbb{R}^{H \times W \times 2} \rightarrow \mathbb{R}^{H \times W \times d}$, where $H$ and $W$ are pixel height and width of $I$ respectively.

The first stream of the network maps the object heightmap $I_\mathrm{obj}$ to a dense object descriptor map $\mu_\mathrm{obj}$. The second stream maps a batch of kit heightmaps $I_\mathrm{kit}$ with 20 different orientations (\ie multiples of 18\degree), to a batch of 20 kit descriptor maps $\mu_\mathrm{kit}$. Each pixel $i\in{I_\mathrm{kit}}$ in the kit heightmap maps to 20 kit descriptors (one for each rotation), but only one of them (the most similar) will match to its corresponding object descriptor in $\mu_\mathrm{obj}$. The index of the rotation with the most similar kit descriptor informs the change in object orientation $\theta$ between the picking and placing, \ie $\theta = \frac{360}{20} \times \argmin_{ij}{\|\mu_\mathrm{kit}^{ij} - \mu_\mathrm{obj}^{i}\|^2_2}$ where $j\in\mathbb{Z}:j\in[1,20]$ and $i\in\mathbb{Z}:i\in[1,H\times W]$. In this way, the matching network not only establishes correspondences between object picking and kit placement locations, but also infers the change in object orientation $\theta$ between the pick and place.

The matching module is trained using a pixel-wise contrastive loss, where for every pair of kit and object heightmaps $(I_\mathrm{kit}, I_\mathrm{obj})$, we sample non-matches from $I_\mathrm{obj}$ and all 20 rotations of $I_\mathrm{kit}$ and matches from $I_\mathrm{obj}$ but only the rotation $j$ of $I_\mathrm{kit}$ corresponding to the ground-truth angle.
The loss function thus encourages descriptors to match solely at the correct rotation of the kit image while non-matches are pushed to be at least a feature distance margin $M$ apart. See the supplemental for additional details.

\mypara{Learning ordered assembly.} Each predicted pixel-wise descriptor from the matching network is conditioned on contextual information available inside of its local receptive field (\eg a descriptor changes based on whether its receptive field sees 0, 1, or more objects already inside the kit). The descriptors thus have the capacity to memorize the sequencing for ordered assembly. As a result, both the matching and place modules \textit{implicitly} enable our system to memorize the sequencing, where the learned order of assembly corresponds to the reversed order of disassembly from data collection.

\subsection{Planner}
\label{sec:method-planning}

The planner is responsible for integrating information from all three modules and producing the final assembly parameters $p$, $q$ and $\theta$. Specifically, top-k pick candidates are sampled from the suction module output $S$ and top-k place candidates are sampled across all 20 rotations of the place module output $P$. Then, for each pick and place pair in the Cartesian product of candidates, kit and object descriptors are indexed and their L2 distance is evaluated after which the pair with the lowest L2 distance across all rotations and all candidates is chosen to produce the final kit descriptor, object descriptor and rotation index.


\section{Automatic Data Collection Via Disassembly\label{sec:datacol}}

\begin{figure}[t!]
\centering

  \includegraphics[width=\linewidth]{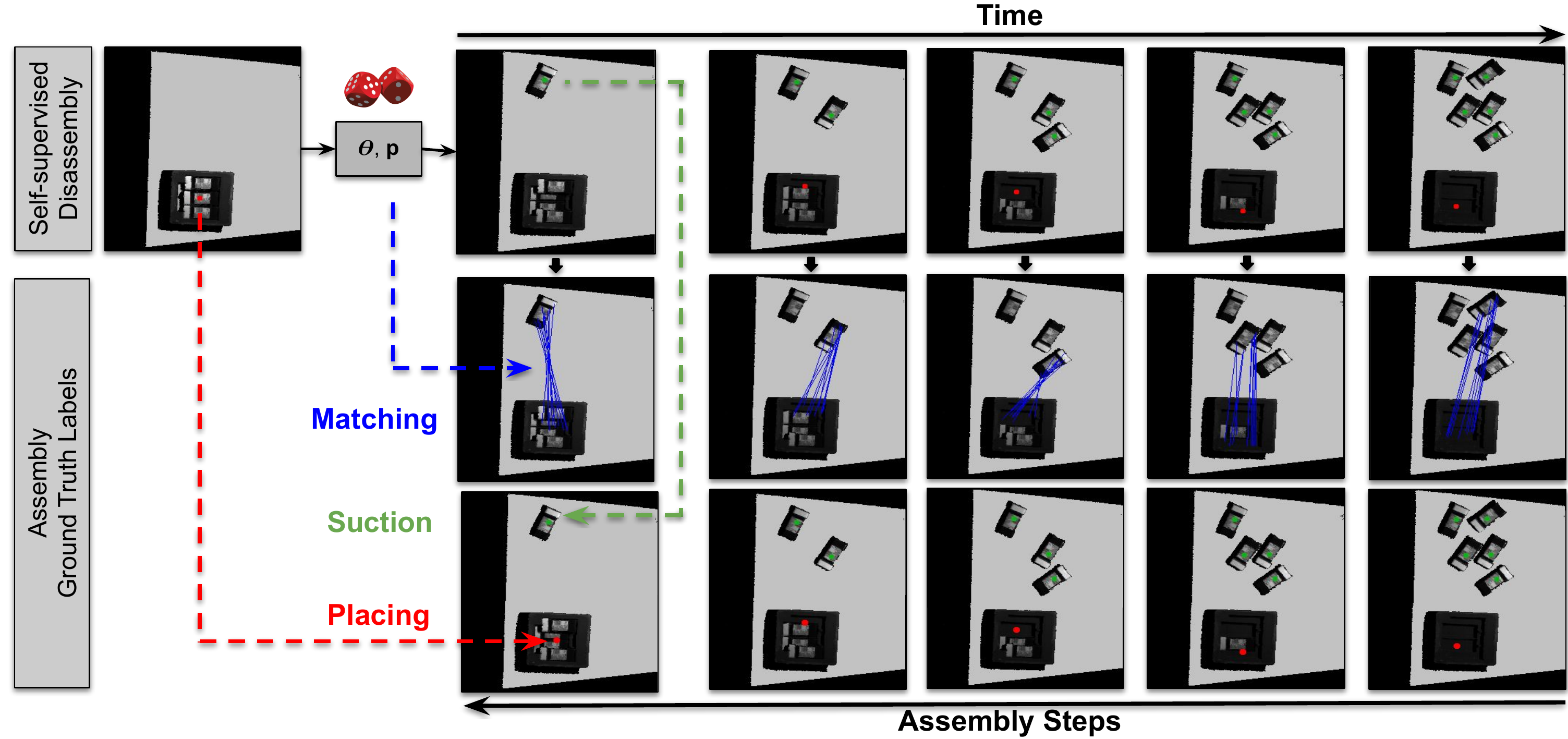}
    \vspace{-2.5mm}
  \caption{\textbf{Self-supervised time reversal} generates ground-truth pick and place correspondences from disassembly data. Concretely, at every timestep, disassembly trajectories are executed by querying the suction network (\ie red) and randomly generating placing poses (\ie green). By capturing the state of the environment before and after the trajectory, the recorded information can be reversed to produce assembly data (\eg the place pose becomes a suction label and the suction pose becomes a place label).
  }
  \label{fig:time-reversal}
  \vspace{-7mm}
\end{figure}

To generate the inputs and ground-truth labels needed to train our various networks, we create a self-resetting closed-loop system wherein the robot continuously generates a random sequence of disassembly trajectories $S = \{T_1, T_2, \ldots , T_N\}$ to empty a kit of $N$ objects, then performs it in reverse $S^{'} = \{T_N, T_{N-1}, \ldots , T_1\}$ to reset the system to its initial state. Fig. \ref{fig:time-reversal} shows the pipeline of the disassembly data collection process.

Specifically, for every object in the kit, a trajectory $T$ is generated as follows: first, the robot captures a grayscale-depth image to construct kit and object heightmaps $I_\mathrm{kit}$ and $I_\mathrm{obj}$, then it performs a forward pass of the suction network to make a prediction of parameter $p$ which is executed by the suction primitive to grasp the object. If the suction action is successful, it places the object at a position $q$ and rotation $\theta$ sampled uniformly at random in the bounds of the workspace. If the suction action is not successful (\ie no object gets picked up), this suction point is labeled as negative for the online learning process.
The suction success signal is obtained by visual background subtraction and measuring suction air flow. To ease trial and error during disassembly, we (a) affix the kit to the workspace to prevent failed grasps from causing accidental displacements and (b) bootstrap the suction network training by manually labelling 50 examples per kit.

All the parameters (\eg $p$, $q$, $\theta$) are stored for resetting the scene. Once all $N$ objects have been disassembled, the robot indexes the trajectories in reverse, suctioning the objects using the place parameters and placing them back into the kit using the suction parameters.
For each trajectory, it stores the grayscale-depth heightmaps captured before and after the disassembly step, the predicted suction pose, and the randomly generated place pose. For each kit in the training set, we collect $\sim 500$ disassembly sequences, which in total takes 8-10 hours.

\mypara{Place Network Dataset.} To generate the training data for the placing network, we use the suction location $p$ at time $t$ and the heightmap $I_\mathrm{kit}$ at time $t+1$ (\ie image taken after the suction action) as one training pair. Thus, the placing network is encouraged to look at empty locations in the kit and predict valid place positions for the next object.

\mypara{Suction Network Dataset.} The training data for the suction network consists of two sets of input-label pairs: (1) the kit heightmap and the suction position $(I_\mathrm{kit}, p)$ and (2) the object heightmap and the place position $(I_\mathrm{obj}, q)$. Thus, the suction network is encouraged to predict favorable suction locations on the object of interest inside \textit{and} outside the kit.

\mypara{Matching Network Dataset.} To label the correspondences for the matching network, we first compute the masks of the object (both inside and outside the kit) using the images' difference. The relationship of every pixel in the cavity of the kit and its corresponding pixel on the object outside the kit is calculated using the rotation angle $\theta$, \ie $[u_{obj}, v_{obj}] = R_{z, \theta} \cdot [u_{kit}, v_{kit}, 1]$ where $R_{z, \theta}$ is a $2 \times 3$ rotation matrix defining a rotation of $\theta \degree$ around the z-axis. Additional details on data collection and pre-training are in the supplemental file.

\begin{figure}[t]
\centering
  \includegraphics[width=\linewidth]{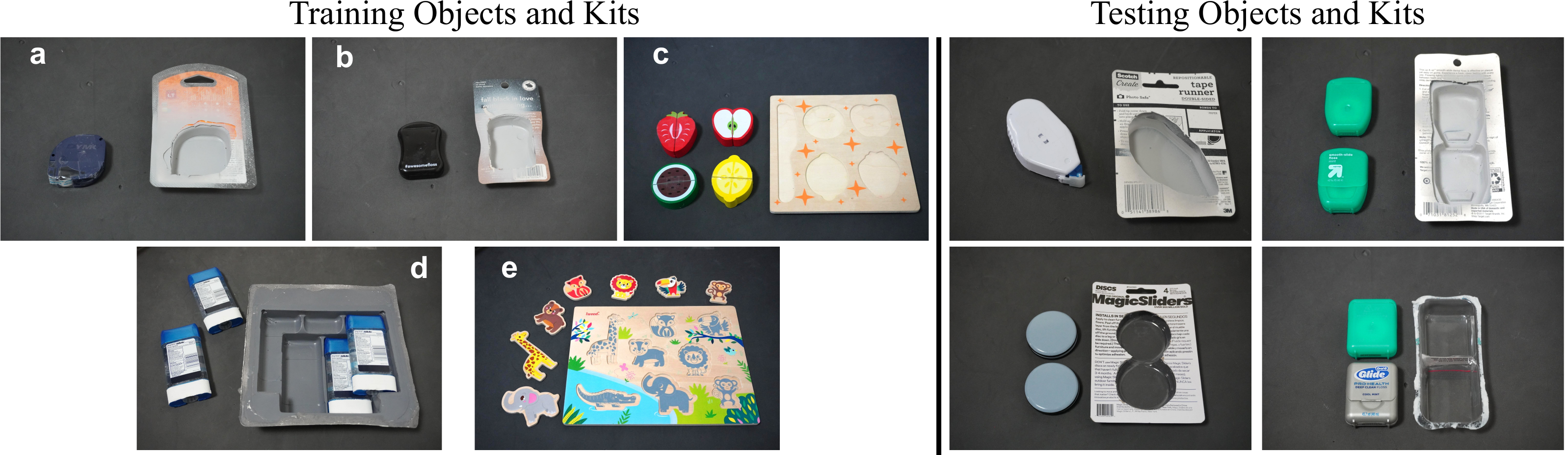}
  \caption{\textbf{Train and Test Distribution}. The training kits: (a) tape-runner, (b) black-floss, (c) fruits, (d) deodorants, and (e) zoo-animals. Two animals are withheld from the zoo-animals kit (i.e. it is only trained on 5 animals) to speed up data collection and testing.} 
  \label{fig:object-distribution}
  \vspace{-5mm}
\end{figure}


\section{Evaluation}

We design a series of experiments to evaluate the effectiveness of our approach across different assembly settings. In particular, our goal is to examine the following questions: (1) How does our proposed method -- based on learned shape-driven descriptors -- compare to other baseline alternatives? (2) How accurate and robust is our system across a wide range of rotations and translations of the objects and kit? (3) Is our system capable of generalizing to new kit configurations such as multiple versions of the same kit and mixtures of different kits when trained solely on individual kits? (4) Does our system learn descriptors that can generalize to previously unseen objects and kits?

\vspace{-1.5mm}
\subsection{Comparison with Pose Estimation Baseline}

Our first experiment compares the kitting performance of Form2Fit to ORB-PE, a classic method for pose estimation that leverages Oriented Fast and Rotated Brief (ORB) descriptors \cite{orb2011} with RANSAC \cite{fischler1981random}. Implementation details for this baseline can be found in the supplementary file.

\mypara{Benchmark.} We collected data from 25 random test sequences for each training kit (shown in  Fig. \ref{fig:object-distribution}) using our automatic data collection pipeline (Sec. \ref{sec:datacol}) and generate associated ground-truth pose transforms (\ie a $4\text{x}4$ matrix encoding the change in object pose from its initial position outside the kit to its final position inside the kit hole).

\mypara{Evaluation metric.} Form2Fit generates a rigid transform between the object and its final position in the kit using the predicted descriptors from the matching module, while ORB-PE generates the rigid transform by matching the observed object to a previously known canonical object model (whose transform into the kit is also known beforehand). Our goal is to evaluate how accurate these rigid transforms compare to the ground truth pose transforms from the benchmark. To this end, we adopt the average distance (ADD) metric proposed in \cite{hinter2012} and measure the area under the accuracy-threshold curve using ADD, where we vary the threshold for the average distance (in meters) and then compute the pose accuracy. The maximum threshold is set to 10 cm. Results are shown in Table \ref{table:main-auc}.

\begin{table}[h]
\vspace{-0.5em}
  \centering
  \setlength{\tabcolsep}{3.5 pt}
  \caption{Area Under The Accuracy-Threshold Curve}
  \vspace{-3mm}
  \begin{tabular}{lccccc}
  \toprule
  Method & Tape-Runner & Black-Floss & Zoo-Animals & Fruits & Mean \\
  \midrule
  ORB-PE & 0.081 & 0.095 & \bf{0.097} & 0.084 &0.089 \\
  Form2Fit & \bf{0.097} & \bf{0.096} & 0.094 & \bf{0.089} & \textbf{0.094} \\
  \bottomrule
  \label{table:main-auc}
  \vspace{-5mm}
  \end{tabular}
  \end{table}

In general, we observe that Form2Fit has a higher mean area under the curve across the different training kits than ORB-PE. While ORB-PE performs competitively, it requires prior object-specific knowledge (\ie canonical object models and their precomputed pose transforms into the kit), making it unable to generalize to novel objects and kits. On the contrary, Form2Fit is capable of generalizing to novel objects and kits, which we demonstrate in the following subsection.

\subsection{Generalization to Novel Settings}

We evaluate the generalization performance of Form2Fit by conducting a series of experiments on a real platform, which consists of a 6DoF UR5e robot using a 3D printed suction end-effector overlooking a tabletop scenario, as well as a Photoneo PhoXi Model M camera, calibrated with respect to the robot base using the pipeline in \cite{zeng2018learning}. Video recordings can be found in our supplementary material.

\mypara{Evaluation metric} is assembly accuracy $\bar{s}$, defined as the percentage of attempts where the objects are successfully placed into their target locations. For the kits that contain multiple objects (\eg deodorants, zoo animals, fruits), we record the average individual success rate of each object, then average over all objects to compute the overall kit success rate, \ie $\bar{s} = \frac{1}{20N} \sum_{n=1}^{N} \sum_{i=1}^{20} s_{ni}$. When evaluating on the training objects, we count 180\degree \ rotational flips as incorrectly assembled even though the objects may still fit in the kit. Our expectation is that the system should pick up on minor details in texture and geometry which should inform it about the right orientation. However, for novel kits in generalization experiments, we choose not to penalize the performance since the system has never previously seen the kits. See the supplemental file for examples of rotational flips.

\begin{table}[t]
\vspace{-1mm}
\caption{Success Rate (Mean) Under Varying Initial Conditions}
\vspace{-3mm}
\centering
\setlength{\tabcolsep}{2pt}
\begin{tabular}{@{}cl|l@{}}
\toprule
Kit Type  &    Name      & Assembly Success \\ \midrule
\multirow{2}{*}{\begin{tabular}[c]{@{}c@{}}single object\\ kits\end{tabular}}& tape-runner & 0.90 \\
& black-floss & 0.95 \\
\multirow{3}{*}{\begin{tabular}[c]{@{}c@{}}multi-object\\ kits\end{tabular}}
& fruits      & 0.65, 0.60, 0.95, 0.90 \\
& zoo-animals & 0.90, 0.95, 0.95, 0.95, 0.85 \\
& deodorants  & 1.00, 1.00, 1.00, 0.90, 0.85 \\
\bottomrule
\label{table:result-single}
\vspace{-3mm}
\end{tabular}

\centering
\caption{Success Rate (Mean) For Novel Configurations}
\vspace{-3mm}
\begin{tabular}{llccc}
\toprule
Config. & 1 black-floss & 1 tape & 3 tape & 2 tape \& 2 black-floss \\ \midrule
Succ. Rate & 0.95 & 0.90 & 1.00, 0.95, 0.85 & 0.80, 1.00, 1.00, 0.95 \\
\bottomrule
\label{table:result-multi}
\end{tabular}
\vspace{-5mm}
\end{table}

\begin{figure}[t]
\centering
\includegraphics[width=\linewidth]{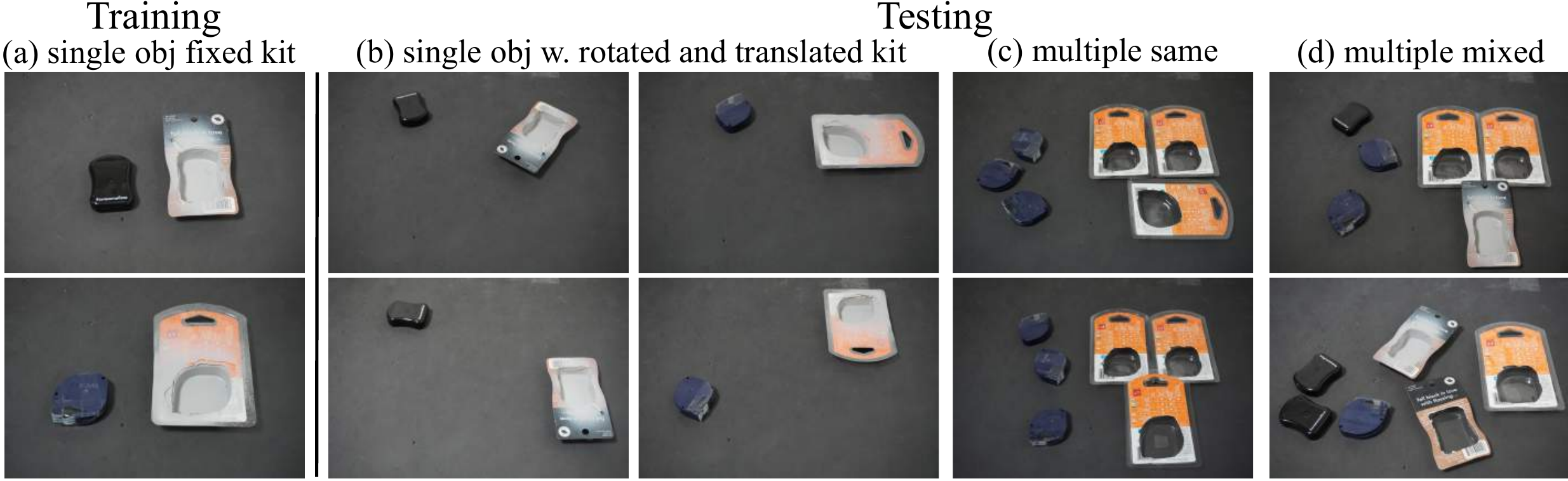}
\caption{\label{fig:setting} Generalization to novel kit configurations.}
\vspace{-7mm}
\end{figure}

\mypara{Generalization to initial conditions.} First, we measure the robustness of our system to varying initial conditions. For each kit, during training, the kit is fixed in the same position and orientation (Fig. \ref{fig:setting} (a)) while during testing, we randomly position and orient it on the workspace (Fig. \ref{fig:setting} (b)). Specifically, we record assembly accuracy on 5 random kit poses, 4 times each for a total of $N = 20$ trials. Kit and object positions are uniformly sampled inside the table, while the orientation of the objects is sampled in $[-180 \degree, 180 \degree]$ and the orientation of the kit in $[-120 \degree, 120 \degree]$. Note that for kits with multiple objects, if the execution of an object fails, we intervene by placing it in its correct location to allow the system to resume. As seen in Table \ref{table:result-single}, our system achieves 90\% average assembly success on both single and multi-object kits. For the fruit kit, our system sometimes mistakes the lemon for the strawberry (and vice versa). We surmise that this is due to their similar geometries in the downscaled heightmap image. In general, we observe that frequent modes of failure come from the robot placing objects (\eg floss and tape) $180 \degree$ flipped from their correct orientation.

\mypara{Generalization to multiple kits.} Next, we study how well our system can generalize to different kit configurations. During training, the system sees only 2 individual kits (Fig. \ref{fig:setting} (a)), while during testing, we create combinations of the same kit and mixtures of kits (Fig. \ref{fig:setting} (c) and (d)). Similarly to above, we perform $5\times4=20$ trials. While our system has never been trained on these novel settings, it is able to achieve an assembly success rate of 94.27\% (Table \ref{table:result-multi}).

\mypara{Generalization to novel kits.} Finally, we study how well our system can generalize to novel objects and kits. Specifically, the testing kits are never-before-seen single-object kits and multi-object kits with various object shapes (see Fig. \ref{fig:object-distribution}). For perfectly symmetrical objects (\eg circles), we consider the assembly to be successful as long as the object is placed into kit, otherwise we count it as a failure. On a set of 20 trials, our system achieves over 86\% generalization accuracy.

\subsection{Feature Visualization}
\begin{figure}[t!]
\centering
  \includegraphics[width=\linewidth]{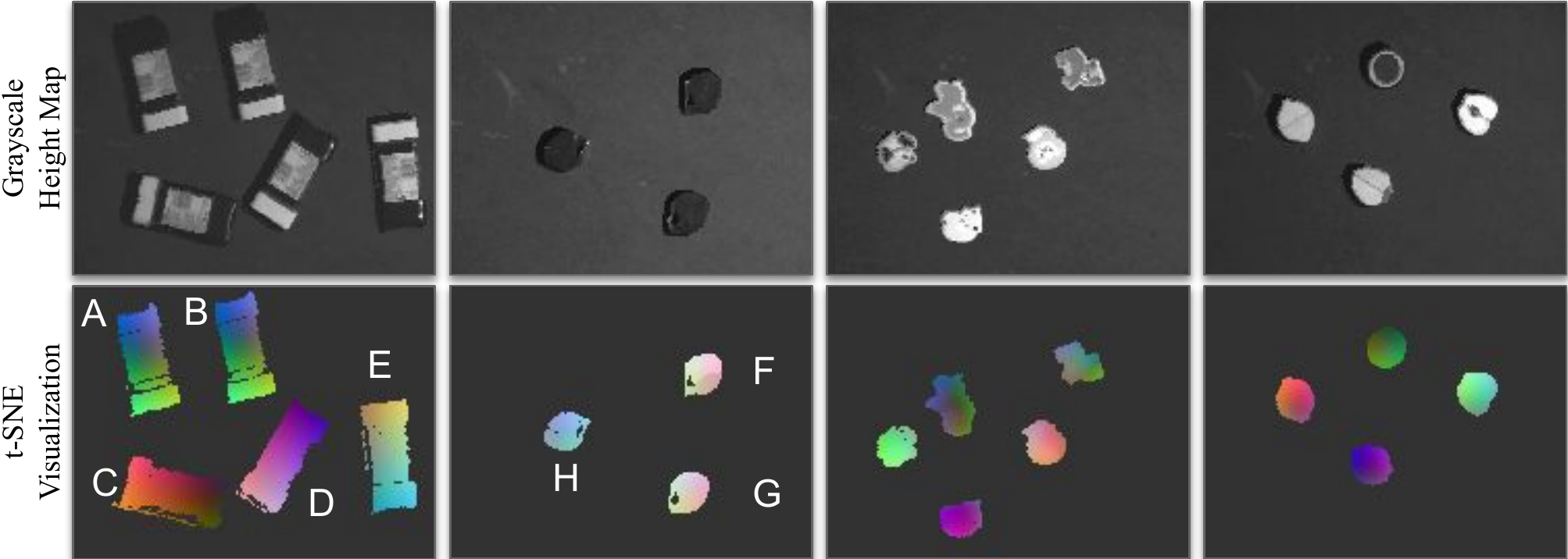}
  \caption{The t-SNE embedding of object descriptors for different kits show that the descriptors have learned to encode rotation (A, B, C), spatial correspondences (identical points on A and B share similar descriptors) and object identity (all zoo animals have unique descriptors).}
  \vspace{-7mm}
  \label{fig:tsne}
\end{figure}

To gain an understanding of the learned object descriptors, we visualize their t-SNE embeddings \cite{van2014accelerating}. Specifically, we mask object pixels in the heightmap using masks obtained in the data collection process and forward them through the matching network. Then, the descriptor map of channel dimension 64 is reduced to dimension 3 using t-SNE and normalized to $[0, 255]$ for color visualization. From Fig. \ref{fig:tsne}, we observe that the descriptors have learned to encode: (a) rotation: objects oriented differently have different descriptors $(A,C,D,E)$ and $(H, F)$, (b) spatial correspondence: same points on the same oriented objects share similar descriptors $(A,B)$ and $(F,G)$, and (c) object identity: zoo animals and fruits exhibit unique descriptors.

\section{Discussion and Future Work}

We present Form2Fit, a framework for generalizable kit assembly. By formulating the assembly task as a shape matching problem, our method learns a general matching function that is robust to a variety of initial conditions, handles new kit combinations, and generalizes to new objects and kits. The system is self-supervised -- obtaining its own training labels for assembly by disassembling kits through trial and error with pick and place, then rewinding the action sequences over time.

However, while our system presents a step towards generalizable kit assembly, it also has a few limitations. First, it only handles 2D rotations (\ie planar object rotations) and assumes that objects are face-down -- it would be interesting to explore a more complex (\eg higher DoF) action representation for 3D assembly. Second, while our system is able to handle partially transparent kits, it has trouble handling fully transparent ones like the deodorant blister pack (we spray-paint it to support stereo matching for our 3D camera). Exploring the use of external vision algorithms like \cite{Xu_2015_transcut,guo2019transparent,chen2018tomnet,ji2017fusing} to estimate the geometry of the transparent kits before using the visual data would be a promising direction for future research.


\newpage
\bibliographystyle{IEEEtran}
\bibliography{IEEEabrv,root}

\end{document}